\documentclass[conference,compsoc,twocolumn]{IEEEtran}

\ifCLASSOPTIONcompsoc
  \usepackage[nocompress]{cite}
\else
  \usepackage{cite}
\fi

\ifCLASSINFOpdf
\else
\fi

\usepackage{amsmath}
\usepackage{algorithmic}
\usepackage{array}
\usepackage{url}
\usepackage{subfig}
\usepackage{graphicx}

\usepackage[utf8]{inputenc}
\usepackage{booktabs}

\begin{document}

\title{Fast and accurate object detection in high resolution 4K and 8K video using GPUs}

\author{
\IEEEauthorblockN{Vít Růžička and Franz Franchetti}
\IEEEauthorblockA{Department of Electrical and Computer Engineering, Carnegie Mellon University \\
Email: previtus@gmail.com, franzf@cmu.edu
}
}
\maketitle


\begin{abstract}
Machine learning has celebrated a lot of achievements on computer vision tasks such as object detection, but the traditionally used models work with relatively low resolution images. The resolution of recording devices is gradually increasing and there is a rising need for new methods of processing high resolution data. We propose an attention pipeline method which uses two staged evaluation of each image or video frame under rough and refined resolution to limit the total number of necessary evaluations. For both stages, we make use of the fast object detection model YOLO v2. We have implemented our model in code, which distributes the work across GPUs. We maintain high accuracy while reaching the average performance of 3-6 fps on 4K video and 2 fps on 8K video.
\end{abstract}

\IEEEpeerreviewmaketitle


\section{Introduction}
\label{1_intoduction}

Machine learning is a fast moving field which has experienced a revolution in working with imagery data for the task of object classification and detection. Typical use of these computer vision tasks include security initiatives like facial recognition, city planning efforts like traffic density estimations \cite{FCN-rLSTM-vehicle-count}. 

Current state of the art in object detection is using deep convolutional neural networks models trained on ImageNet dataset \cite{imagenet} such as Faster R-CNN \cite{faster_rcnn} and YOLO \cite{yolo_v2}. 

The majority of these models are focused on working with low-resolution images for these three following reasons. First, in certain scenarios, the low-resolutions images are sufficient for the task, such as in the case of object classification, where most models use images up to 299x299 pixels \cite{alexnet2012,vgg16,resnet,inception}. Secondly, processing low-resolution images is more time efficient. Lastly, many public available datasets used to train these models such as ImageNet, CIFAR100 \cite{cifar100}, Caltech 256 \cite{caltech256} and LFW \cite{lfw} are themselves made up of low-resolution images. There are no large scale datasets with more than hundred of images or videos and resolution as high as 4K (3840x2160 pixels).

However, in low resolution images one can lose a lot of detail that is not forfeited when using high resolution capture devices. Today’s high resolution data sources introduce 4K-8K cameras therefore bringing a need for new models or methods to analyze them. Also, there are advantages in how much information we can extract from higher resolution images.  For example in Figure \ref{fig:example_annotated} we can detect more human figures in the original resolution as compared to resizing the image to the lower resolution of the models.

With the limitations of current models, we came up with two baseline approaches. First, downscaling the image before evaluation and sacrificing accuracy. Or secondly, cutting up the whole image into overlapping crops and evaluating every single crop while sacrificing speed.


    \smallbreak
    \textbf{Contributions}

In this paper we propose a method for accurate and fast object detection which we call attention pipeline. Our method uses the first approach by downscaling the original image into a low-resolution space. Object detection in low resolution guides the attention of the model to the important areas of the original image. In the second stage, the model is directed back to the high resolution only reviewing the highlighted areas. 

Specifically, in this paper we make the following contributions:

	\begin{figure}[t]
		\centering
		\includegraphics[width=0.45\textwidth]{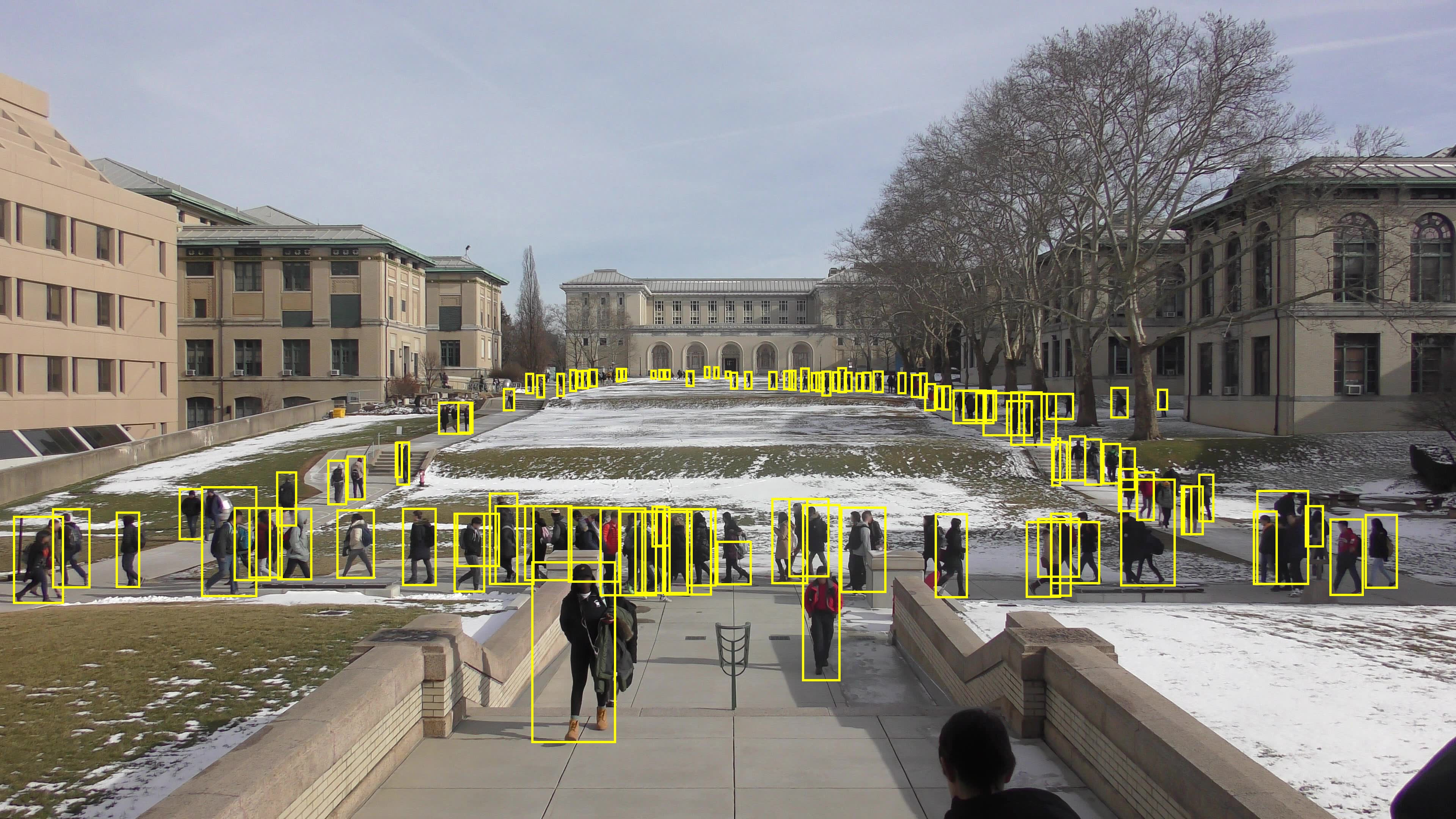}
		\caption{Example of crowded 4K video frame annotated with our method.}
		\label{fig:example_annotated}
	\end{figure}

    \begin{itemize}
    \item We propose a novel method which processes high resolution video data while balancing the trade-off of accuracy and performance.
    \item We show that our method reduces the number of inspected crops as compared to a baseline method of processing all crops in each image and as a result increases performance by up to 27\%.
    \item We increase the PASCAL VOC Average Precision score on our dataset from $33.6 AP_{50}$ to $75.4 AP_{50}$ as compared to using YOLO v2 in baseline approach of downsampling images to the model's resolution.
    \item Implement efficient code which distributes work across GPU cluster and measure the performance of each individual operation of proposed method.
    \end{itemize}


\subsection{Related Work}



	In this section, we will trace the important advancements in the field of machine learning relevant for the task of object detection and the efforts to speed up existing models by using the concept of attention.

\smallbreak
\textbf{Revolution of deep convolutional neural networks.} The success of deep convolutional neural networks (CNN) on imagery data has been initiated by the large annotated dataset of ImageNet and by the ILSVRC competition.  In the ILSVRC 2012 competition, AlexNet model \cite{alexnet2012} has became the new state of the art on object classification while in  PASCAL VOC 2012 and the ILSVRC 2013 challenge the R-CNN model \cite{rcnn} extended the usage of CNNs on the object detection task. 


\smallbreak
\textbf{Efforts to speed up object detection.} One of the approaches for object detection depends on hierarchy of region proposal methods \cite{RecognitionUsingRegions}. The initial work of R-CNN uses CNNs to classify objects in the proposed regions as well as the following Fast R-CNN \cite{fast_rcnn} and Faster R-CNN \cite{faster_rcnn}. Comparative study of \cite{speed_accuracy_tradeoff} explores the performance gains of these models.

	In our paper we are using the YOLO model first introduced in \cite{yolo_v1} and later improved as YOLO v2 in \cite{yolo_v2}. These models have achieved real time performance by unifying the whole architecture into one single neural network which directly predicts the location of regions and their classes during inference. 

	Closest to our approach is the work of \cite{face_4kvid}, which uses region proposal hierarchy on the task of real time face detection in 4K videos. Our method differs in that it generalizes to other classes of objects given by the generality of the YOLO v2 model.

\smallbreak
\textbf{Attention.} The concept of attention, to focus just on few areas of the original image, can be used for two goals. First goal is to increase the model's accuracy such as in the works of \cite{localization-for-classification, parts-attention-finegrained} which tries to select few parts of the image relevant for the task. 

	Second goal is aimed to limit the computational costs. Inspired by human foveal vision, the work of \cite{videoClassifFovea} examines videos under multiple resolutions, using only the center of the image in its original resolution. In our proposed pipeline we will also use subsections of the original image, however our focus will be guided by initial fast yet imprecise attention evaluation.


\section{Method}

In this section we describe our proposed method we call attention pipeline.

\subsection{Problem definition}
\label{subsec-problem_definition}

	In the context of this paper, we will be working with high resolution dataset of videos, which can be seen as a stream of consecutive frames. Additional to the spacial information within each frame, there is temporal information \cite{videoClassifFovea} carried across neighboring frames (the same tracked object might be present in the next frame at similar location).
	


	The task of object detection consists of finding objects of interest in the image by marking them with bounding boxes. Bounding box is a four coordinate rectangle with a class label, which should contain the corresponding object as tightly as possible. One image can contain many possibly overlapping bounding boxes of multiple classes (such as “person”, “car”, etc.).
	
	\begin{figure}[h]
		\centering
		\includegraphics[width=0.45\textwidth]{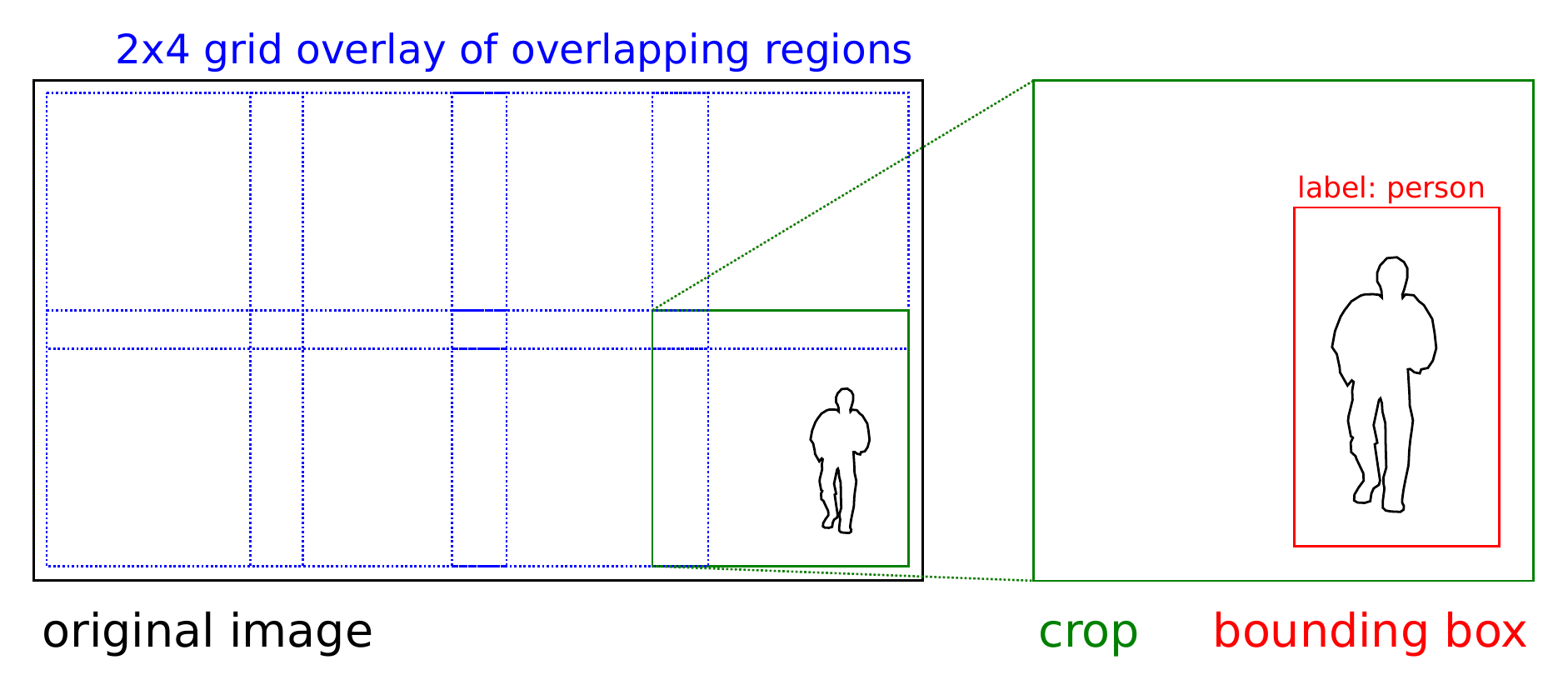}
		\caption{Illustration of the object detection task and terms used in this paper. By cutting out and resizing crop of the original image we can use YOLO v2 model for object detection of objects such as people.}
		\label{fig:intro-figure}
	\end{figure}

	We define the term of “crop” when talking about any subregion of the image. We can overlay any image by overlapping regions and cut out crops of smaller size. See Figure \ref{fig:intro-figure}. We will be using the YOLO v2 model, which is by design limited to square ratio input with fixed resolution of 608x608 pixels. We comply with these limitations for the fast performance YOLO model offers. 

    \smallbreak
    \textbf{Baseline approaches.} There are two baseline solutions for this task which provide us with an inspiration for our proposed pipeline and which we will use as poins of comparison in measuring accuracy and performance. First approach is to downscale the whole image original image into resolution of the evaluating model. This approach offers fast evaluation, but loses large amounts of information potentially hidden in the image, which especially applies with high resolutions. Second approach is to overlay the original image with a slightly overlapping grid of fixed resolution and evaluate each cut out crop separately. In this second approach we pay the full computational cost as we evaluate every single crop of the image.


		\begin{figure}[h]
			\centering
			\includegraphics[width=0.45\textwidth]{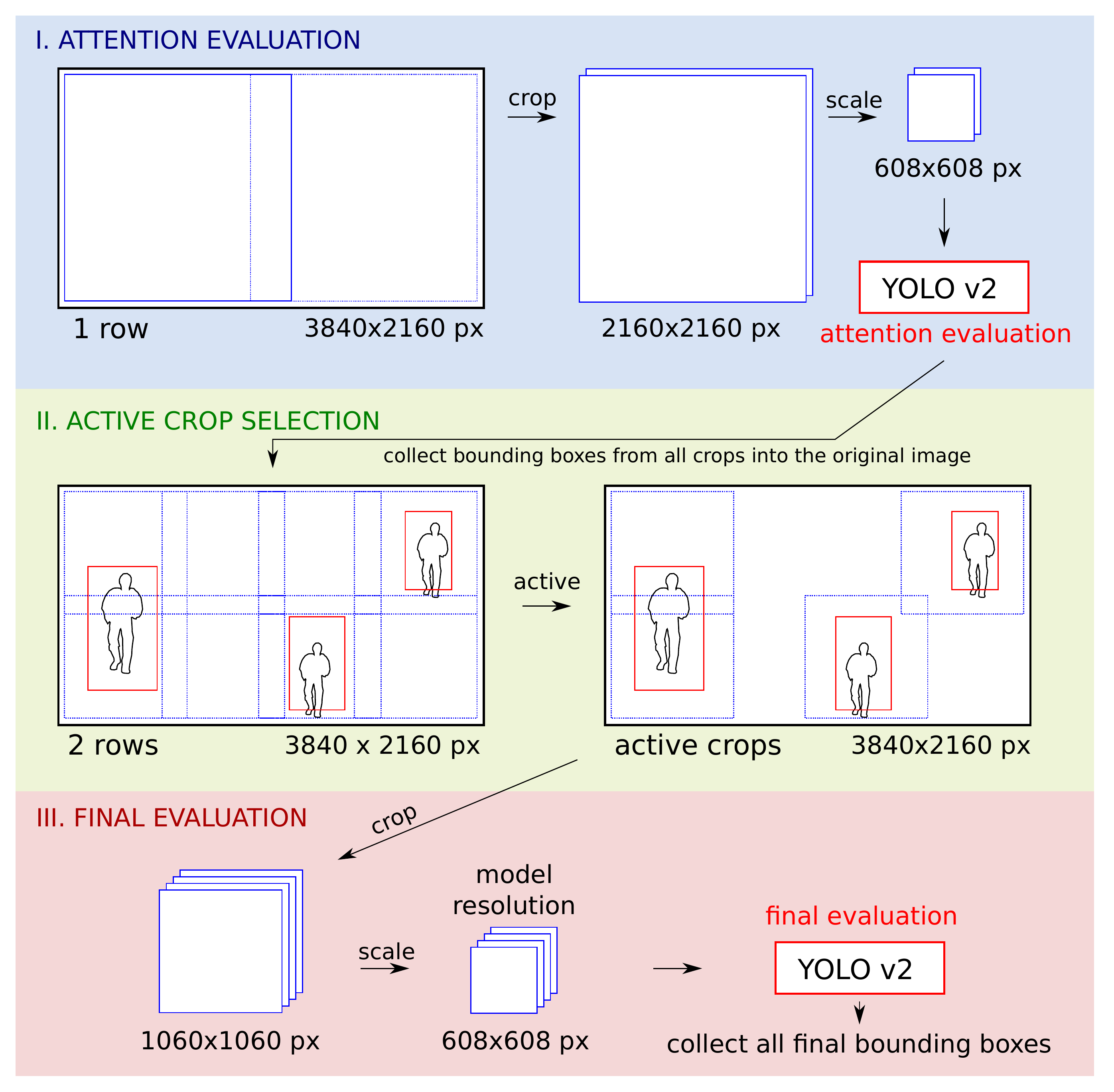}
			\caption{Resolution handling on the example of 4K video frame processing. During the attention step we process the image under rough resolution which allows us to decide which regions of the image should be active in final finer evaluation.}
			\label{fig:method_pipeline_resolution}
		\end{figure}

\subsection{Attention pipeline} \label{sec:method_attention_pipeline}

	We propose an attention pipeline model, which leverages these two basic approaches in striving both for precision and performance. We evaluate the image in staged manner. First “attention evaluation” stage looks at roughly sampled areas of the image to get areas suspicious of presence of objects we are localizing. The second “final evaluation” stage then looks at these selected areas under higher resolution. 

	\begin{figure*}[h]
		\centering
		\includegraphics[width=0.9\textwidth]{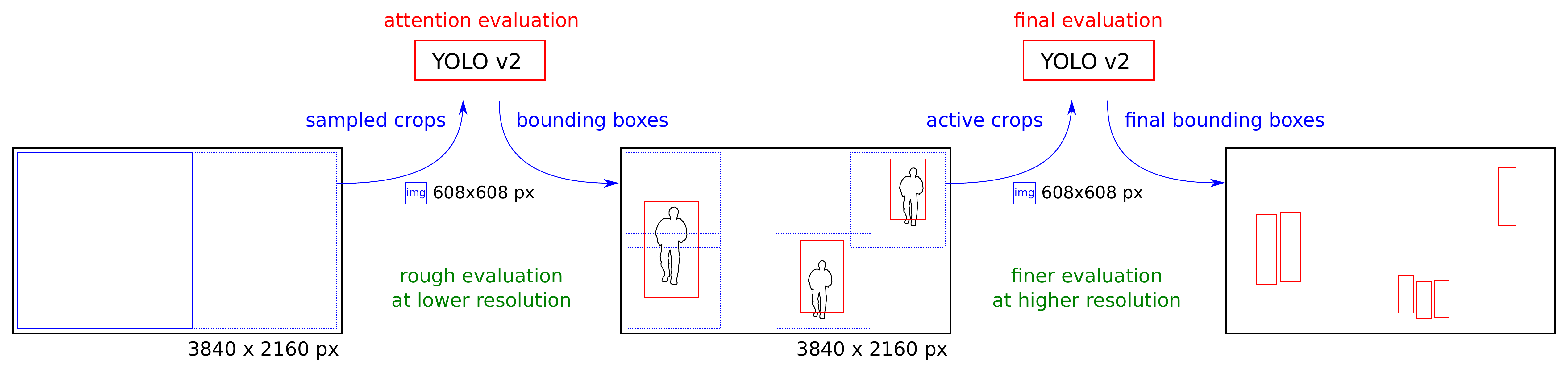}
		\caption{The attention pipeline. Stepwise breakdown of the original image under different effective resolution.}
		\label{fig:LABEL1}
	\end{figure*}


    \smallbreak
    \textbf{Attention evaluation.} \label{sec:subsec_attention_eval_stage}	The original image first enters the attention evaluation stage. For simple way to balance the accuracy and speed we chose to parameterize the cropping by number of rows a grid imposed over the original image will have and by the overlap in pixels between neighboring cells. For example we can choose the attention evaluation to crop the image by a grid of one row and 0 pixel overlap. The image will then be downscaled to have 608px height and width corresponding to the aspect ratio. The width of the image will be subdivided into square crops of 608x608 pixels, such that the original image is fully covered with the minimal amount of squares. Note that these crops can be overlapping. See examples of this grid in stages one and two in Figure \ref{fig:method_pipeline_resolution}.
		
		
		We evaluate these initial attention crops with YOLO v2 model and get bounding boxes of detected objects. Note that this initial evaluation might lose some of the small or occluded objects in the image, however it will still pick up on rough areas of interest. In practical setting of video analysis, we use the temporal aspect of the video and merge and reuse attention across few neighboring frames.
	    
	\smallbreak
	\textbf{Active crop selection.} Secondly, we will subdivide the original image into a finer grid. Each cell of the grid is then checked for intersections with the bounding boxes detected in the attention evaluation. Intersecting cells will be marked as active crops for the final evaluation. The number of active crops is usually lower than the number of all possible crops, depending on the density of the video. 

    \smallbreak
	\textbf{Final evaluation.} Lastly, in the final evaluation, we use the same YOLO v2 model to locate objects in these higher resolution square crops. See stage three in Figure  \ref{fig:method_pipeline_resolution} and note the difference in resolution of crops (1060 px scaled to 608 px instead of the full height of 2160 px scaled to 608 px).

\begin{figure}[h]
			\centering
			\includegraphics[width=0.25\textwidth]{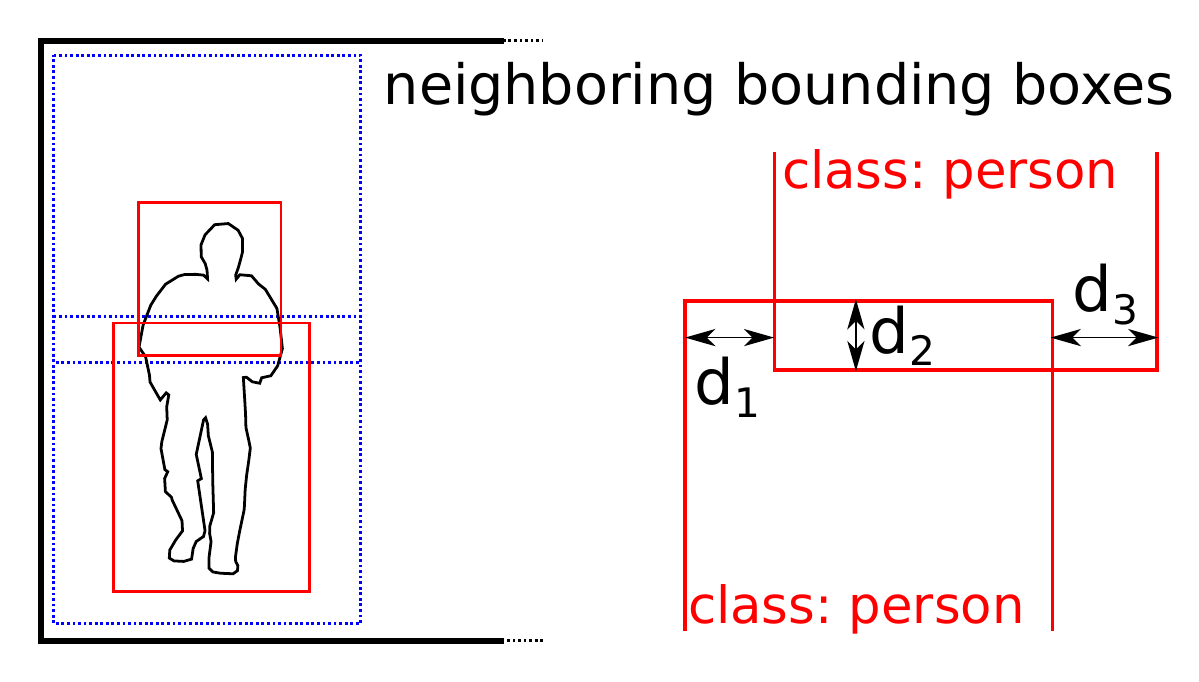}
			\caption{Illustration of object residing on the border of two crop regions. Nearby bounding boxes can be merged in postprocessing step if they are closer than several thresholds.}
			\label{fig:postprocessing}
		\end{figure}

    \smallbreak
	\textbf{Postprocessing.} Upon evaluation of multiple overlapping crop regions of the image, we obtain list of bounding boxes for each of these regions. To limit the number of bounding boxes, we run a non-maximum suppression algorithm to keep only the best predictions. As we do not apriori know the size of the object or its position in the image, we might detect the same object in multiple neighboring regions. This occurs either if the object is larger than the region of our grid, or if it resides on the border of two neighboring regions effectively being cut in half. As is illustrated in Figure \ref{fig:postprocessing}.
	
		Object being cut by the overlaid grid can be detected if we look along the splitting borders. There we can try to detect and merge nearby bounding boxes. This problem is data specific, if we are trying to localize object such as humans, they tend to be high rather than wide in the image. This can be adjusted depending on the detected object class. Empirically we have set several distance thresholds under which we can merge nearby bounding boxes of the same object class. For human detection we consider merging regions neighboring only vertically.


\subsection{Implementation of the client-server version} \label{sec:client_server}

	The motivation of the proposed attention pipeline \ref{sec:method_attention_pipeline} is to make real time evaluation of 4K videos feasible. The following two specific properties of our problem can be leveraged to achieve fast processing.

	The first point is that the image crop evaluation is an embarrassingly parallel problem and as such it lends itself for parallel distribution across multiple workers. 
	
\begin{figure}[h]
		\centering
		\includegraphics[width=0.47\textwidth]{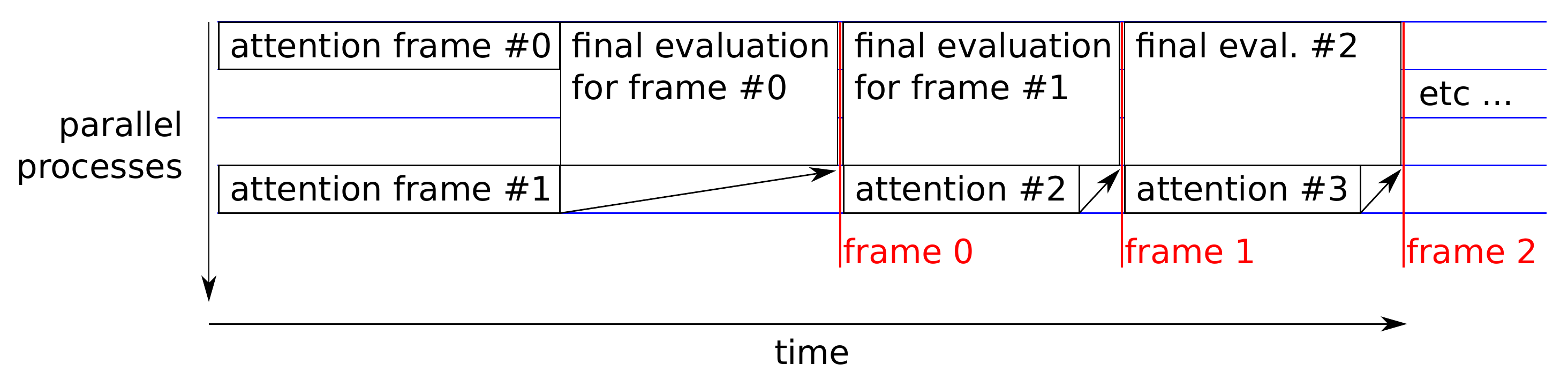}
		\caption{The final evaluation step dependency on the previous attention evaluation step can be decomposed in pipelining manner similar to CPU instruction evaluation.}
		\label{fig:client_server_att_eval_pipe}
	\end{figure}

	Second suggested point is the specific property of our pipeline, where the final evaluation always depends on the previous attention evaluation step. We cannot bypass this dependency, however, we can compute the next frame’s attention evaluation step concurrently with the current frame’s final evaluation. With enough resources and workers, we are effectively minimizing the waiting time for each attention evaluation step. See Figure \ref{fig:client_server_att_eval_pipe}.
	
\begin{figure}[h]
		\centering
		\includegraphics[width=0.45\textwidth]{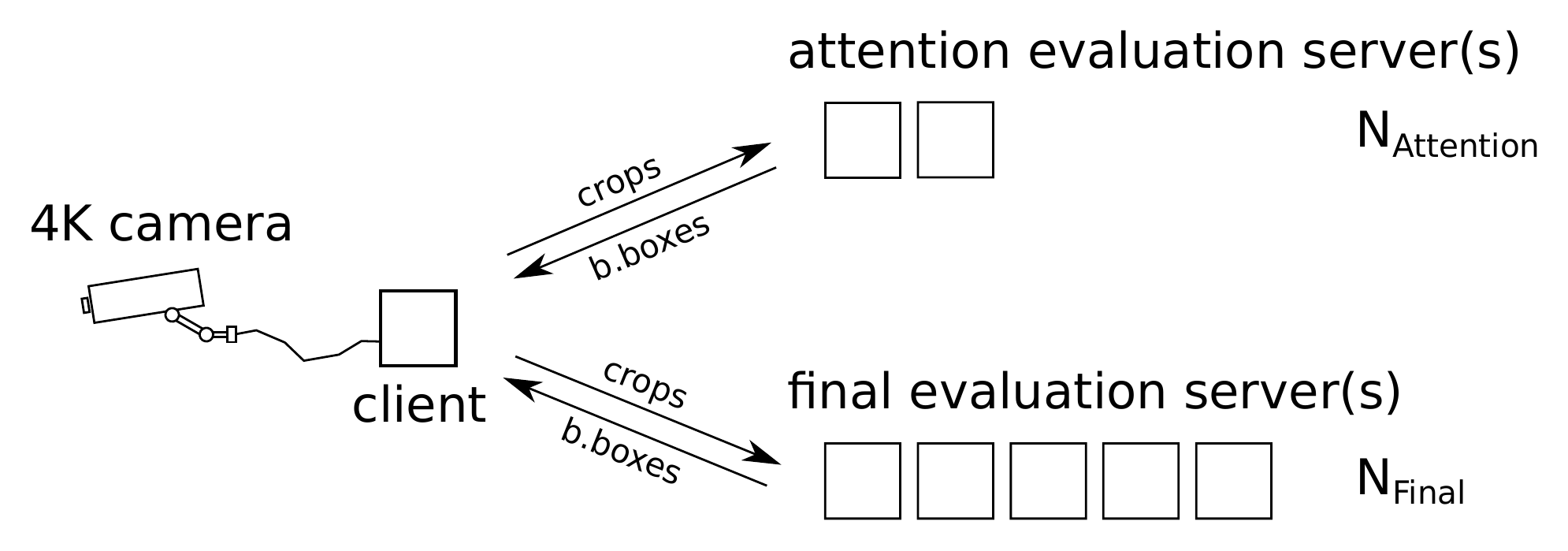}
		\caption{
			Client-server implementation scheme. Client processes the captured video frames and sends only the list of crops for evaluation to the servers. Note that we have dedicated $N_A$ servers for precomputing the next frame’s attention. The $N_F$ servers will have uniformly distributed load of crops to process.
		}
		\label{fig:client_server_attention_final_servers}
	\end{figure}

To leverage these properties, we used a client-server implementation\footnote{Code available at \url{https://github.com/previtus/AttentionPipeline}} as illustrated by Figure \ref{fig:client_server_attention_final_servers}. Note that with strong client machine we can also move the input/output operations and image processing into multiple threads, further speeding up the per frame performance.

\section{Results}

In this section we will first start with the system details, datasets and used metrics and then show the measured accuracy and performance results.

\subsection{Methodology}

\textbf{System details.} We are running our implementation on nodes of the PSC’s Bridges cluster\footnote{\url{https://www.psc.edu/bridges/}}. Each node is equipped with 2 Intel Broadwell E5-2683 v4 CPUs running at 2.1 to 3.0 GHz and 2 NVIDIA Tesla P100 Pascal GPUs. Each node was used by two running instances of code, each utilizing one of the CPUs and one of the GPUs. Peak performance of P100 GPU operating on 32-bit floats is 9.3 TFLOPs. Peak performance of the Intel Broadwell CPU is 768 GFLOPs. Peak bandwidth of transferring data between CPU and GPU was 32 GB/s (PCIe3 x16), while transferring data between nodes was 12.5 GB/s/direction. The worker nodes do not communicate between each other. 


\textbf{Datasets.} We are working with the PEViD-UHD dataset \cite{PEViD-UHD} of 26 short videos of security surveillance scenarios. These contain small number of participants performing labeled actions such as exchanging bags, stealing and other. Some of the 13 seconds long scenes contained not annotated human figures, which is why we have chosen to make a subselection of the PEViD dataset which will be referred to as “PEViD clean” in our measurements. We have removed frames where an unannotated figure was present in the video. For comparability of the results we have kept “PEViD full” untouched. In both cases we chose videos marked as “Exchange” and “Stealing”.

We note that the density of human figures across the PEViD dataset is relatively low (usually just two individuals), which is why we recorded our own dataset using a 4K camera. These include scenes of variety of lighting conditions, distances from the subjects and density of the present human figures. We have also manually labeled a representative sample of ten frames per video with bounding boxes around humans. Full videos are approximately one minute long.

Finally, we have also included an unannotated, publicly available 8K video from YouTube\footnote{\url{https://youtu.be/gdnHLE_HCX0}}.

\textbf{Accuracy metrics.} For accuracy measurement we chose the traditionally used Intersection over Union metric with thresholds 0.5, 0.25 and 0.75. We then analyze the detected true and false positives in an PASCAL VOC average precision score \cite{PASCAL_VOC_AP}, which we refer to as $AP_{50}$, $AP_{25}$ and $AP_{75}$ depending on the used threshold.	

\textbf{Performance metrics.} For performance measurements we use the full version of our own 4K videos, two full videos from the PEViD dataset and the publicly available 8K video.

In our results, we chose to differentiate between stages of evaluation as described in \ref{sec:method_attention_pipeline}. We have separated time measurements for I/O loading and saving, attention evaluation stage as one value and final evaluation stage in more detail. The final evaluation stage is divided into client side image processing, time to transfer images between client and the workers and finally the evaluation of the object detection model itself. As the work is being distributed across multiple servers, and the processing speed is limited by the slowest worker, we chose to show the measurements from the slowest worker per each frame.

\textbf{Crop settings.} Choice of crop settings influences by how many rows we subdivide the original image. We cover the area of image by minimal amount of square crops. Each square is scaled to the model's resolution which is 608x608 pixels. Depending on the settings value, we influence how much we are downscaling the original image.  While originally large objects will likely be detected even after the downscaling, smaller objects in the background might be lost. Table \ref{crop_setting_table} contains the crop sizes in pixels depending on each crop setting. These values include the default 20 pixels overlap between crops.


\if false
\begin{table}[]
\centering
\caption{Crop setting table, sizes of each crop in px}
\label{crop_setting_table}
\begin{tabular}{c|c|c|c|c|}
\cline{2-5}
crop setting & \multicolumn{2}{c|}{4K, 3840x2160 px} & \multicolumn{2}{c|}{8K, 7680x4320 px} \\ \cline{2-5} 
1 row x 2 columns           & 2160 px           & 1x2 grid           & 4320px           & 2 columns           \\ \cline{2-5} 
2            & 1098px           & 2x4 grid           & 2196px           & 2x4 grid           \\ \cline{2-5} 
3            & 736px            & 3x6 grid           & 1472px           & 3x6 grid           \\ \cline{2-5} 
4            & 554px            & 4x8 grid           & 1107px            & 4x8 grid           \\ \cline{2-5} 
6            & 370px            & 6x11 grid          & 480px            & 6x11 grid          \\ \cline{2-5} 
\end{tabular}
\end{table}
\fi

\begin{table}[]
\centering
\caption{Crop setting table, sizes of each crop in px}
\label{crop_setting_table}
\begin{tabular}{@{}lrrrrr@{}}
\toprule
\textbf{} & \textbf{1x2 grid} & \textbf{2x4 grid} & \textbf{3x6 grid} & \textbf{4x8 grid} & \textbf{6x11 grid} \\ \midrule
4K        & 2160 px            & 1098 px            & 736 px             & 554 px             & 370 px              \\
8K        & 4320 px            & 2196 px            & 1472 px            & 1107 px            & 480 px              \\ \bottomrule
\end{tabular}
\end{table}


\subsection{Accuracy analysis}

\if false
\begin{table*}[]
\centering
\caption{Accuracy}
\label{fig:3_1_table_accuracy}
\begin{tabular}{@{}lccccccc@{}}
\toprule
\textbf{dataset name}         & \multicolumn{1}{l}{\textbf{resolution}} & \multicolumn{1}{l}{\textbf{\# of videos}} & \multicolumn{1}{l}{\textbf{\# of frames}} & \multicolumn{1}{l}{\textbf{Settings}} & \multicolumn{1}{l}{\textbf{AP .75}} & \multicolumn{1}{l}{\textbf{AP .50}} & \multicolumn{1}{l}{\textbf{AP .25}} \\ \midrule
PEViD "exchange, steal" clean & 4K                                      & 10                                        & 2992                                      & downscale baseline                      & 0.299                               & 0.657                               & 0.749                               \\
                              &                                         &                                           &                                           & 1 att, 2 fin, 50 over                 & 0.179                               & 0.731                               & 0.838                               \\
                              &                                         &                                           &                                           & 1 att, 3 fin, 50 over                 & \textbf{0.434}                      & \textbf{0.917}                      & \textbf{0.943}                      \\ \midrule
PEViD "exchange, steal" full  & 4K                                      & 13                                        & 4784                                      & downscale baseline                      & 0.274                               & 0.651                               & 0.729                               \\
                              &                                         &                                           &                                           & 1 att, 2 fin, 50 over                 & 0.190                               & 0.713                               & 0.819                               \\
                              &                                         &                                           &                                           & 1 att, 3 fin, 50 over                 & \textbf{0.431}                      & \textbf{0.887}                      & \textbf{0.924}                      \\ \midrule
densely populated scenes      & 4K                                      & 11                                        & 112                                       & downscale baseline                      & 0.070                               & 0.330                               & 0.443                               \\
                              &                                         &                                           &                                           & 1 att, 2 fin, 20 over                 & 0.273                               & 0.610                               & 0.655                               \\
                              &                                         &                                           &                                           & 1 att, 3 fin, 20 over                 & 0.284                               & 0.633                               & 0.703                               \\
                              &                                         &                                           &                                           & 2 att, 4 fin, 20 over                 & \textbf{0.371}                      & \textbf{0.743}                      & \textbf{0.796}                      \\
                              &                                         &                                           &                                           & 2 att, 6 fin, 20 over                 & 0.348                               & 0.663                               & 0.765                               \\ \bottomrule
\end{tabular}
\end{table*}

\begin{table*}[]
\centering
\caption{Accuracy}
\label{fig:3_1_table_accuracy}
\begin{tabular}{lccclccc}
\hline
\textbf{dataset name}         & \textbf{resolution} & \textbf{\# of videos} & \textbf{\# of frames} & \textbf{Settings}         & \textbf{AP .75} & \textbf{AP .50} & \textbf{AP .25} \\ \hline
PEViD "exchange, steal" clean & 4K                  & 10                    & 2992                  & downscale baseline        & 0.321           & 0.739           & 0.843           \\
                              &                     &                       &                       & 1 att, 2 fin, 50 over     & 0.179           & 0.731           & 0.838           \\
                              &                     &                       &                       & 1 att, 3 fin, 50 over     & \textbf{0.434}  & \textbf{0.917}  & \textbf{0.943}  \\
                              &                     &                       &                       & 2 att, 4 fin, 50 over     & 0.383           & 0.794           & 0.911           \\
                              &                     &                       &                       & all crops baseline, 3 fin & \textbf{0.435}  & \textbf{0.938}  & \textbf{0.947}  \\ \hline
PEViD "exchange, steal" full  & 4K                  & 13                    & 4784                  & downscale baseline        & 0.29            & 0.742           & 0.834           \\
                              &                     &                       &                       & 1 att, 2 fin, 50 over     & 0.19            & 0.713           & 0.819           \\
                              &                     &                       &                       & 1 att, 3 fin, 50 over     & \textbf{0.431}  & \textbf{0.887}  & \textbf{0.924}  \\
                              &                     &                       &                       & 2 att, 4 fin, 50 over     & 0.398           & 0.797           & 0.898           \\
                              &                     &                       &                       & all crops baseline, 3 fin & \textbf{0.431}  & \textbf{0.907}  & \textbf{0.949}  \\ \hline
densely populated scenes      & 4K                  & 11                    & 112                   & downscale baseline        & 0.071           & 0.336           & 0.448           \\
                              &                     &                       &                       & 1 att, 2 fin, 20 over     & 0.273           & 0.61            & 0.655           \\
                              &                     &                       &                       & 1 att, 3 fin, 20 over     & 0.284           & 0.633           & 0.703           \\
                              &                     &                       &                       & 2 att, 4 fin, 20 over     & \textbf{0.371}  & \textbf{0.743}  & \textbf{0.796}  \\
                              &                     &                       &                       & all crops baseline, 3 fin & \textbf{0.372}  & \textbf{0.754}  & \textbf{0.812}  \\ \hline
\end{tabular}
\end{table*}
\fi

\begin{table*}[]
\centering
\caption{Accuracy}
\label{fig:3_1_table_accuracy}
\begin{tabular}{@{}lccclccc@{}}
\toprule
\textbf{Dataset name}         & \textbf{Resolution} & \textbf{\# of videos} & \textbf{\# of frames} & \textbf{Settings}         & \textbf{AP$_{75}$} & \textbf{AP$_{50}$} & \textbf{AP$_{25}$} \\ \midrule
PEViD "exchange, steal" clean & 4K                  & 10                    & 2992                  & downscale baseline        & 32.1               & 73.9               & 84.3               \\
                              &                     &                       &                       & 1 att, 2 fin, 50 over     & 17.9               & 73.1               & 83.8               \\
                              &                     &                       &                       & 1 att, 3 fin, 50 over     & \textbf{43.4}      & \textbf{91.7}      & \textbf{94.3}      \\
                              &                     &                       &                       & all crops baseline, 3 fin & \textbf{43.5}      & \textbf{93.8}      & \textbf{94.7}      \\ \midrule
PEViD "exchange, steal" full  & 4K                  & 13                    & 4784                  & downscale baseline        & 29.0               & 74.2               & 83.4               \\
                              &                     &                       &                       & 1 att, 2 fin, 50 over     & 19.0               & 71.3               & 81.9               \\
                              &                     &                       &                       & 1 att, 3 fin, 50 over     & \textbf{43.1}      & \textbf{88.7}      & \textbf{92.4}      \\
                              &                     &                       &                       & all crops baseline, 3 fin & \textbf{43.1}      & \textbf{90.7}      & \textbf{94.9}      \\ \midrule
densely populated scenes      & 4K                  & 11                    & 112                   & downscale baseline        & 7.1                & 33.6               & 44.8               \\
                              &                     &                       &                       & 1 att, 2 fin, 20 over     & 27.3               & 61.0               & 65.5               \\
                              &                     &                       &                       & 2 att, 4 fin, 20 over     & \textbf{37.1}      & \textbf{74.3}      & \textbf{79.6}      \\
                              &                     &                       &                       & all crops baseline, 4 fin & \textbf{37.2}      & \textbf{75.4}      & \textbf{81.2}      \\ \bottomrule
\end{tabular}
\end{table*}


	We report the results of several settings used with each video set in Table \ref{fig:3_1_table_accuracy}. We will use the naming scheme of combining the crop settings used during attention and final evaluation. 

	\begin{figure}[h]
		\centering
		\includegraphics[width=0.41\textwidth]{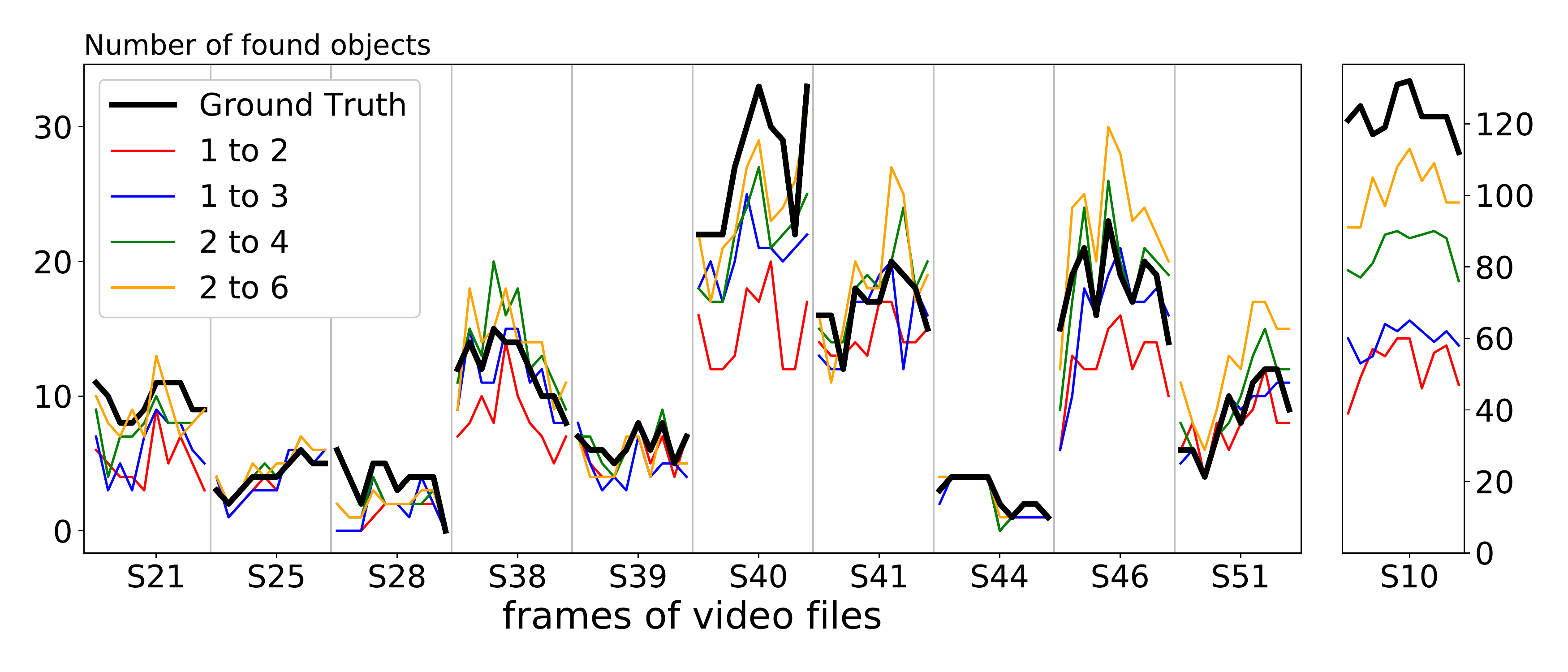}
		\caption{Number of detected object under different settings as compared with the annotated ground truth. Note that video S10 has been plotted separately as it contains vastly larger amount of human figures present.}
		\label{fig:3_1_num_object_detected}
	\end{figure}

	Upon inspecting the results on Table \ref{fig:3_1_table_accuracy} we note that our method achieves accuracy of $91.7 AP_{50}$ on the PEViD dataset and accuracy of $74.3 AP_{50}$ on our own recorded densely populated scenes.
	
	We compare our results with two baseline approaches introduced in \ref{subsec-problem_definition}. We refer to a baseline approach which downscales the original 4K image to the resolution of YOLO v2 model as the "downscale baseline". "All crops baseline" denotes the second baseline approach which cuts up the original image and evaluates all resulting crops. Notice that with correct crop setting, our method vastly outperforms the downscale method, while it achieves results close to the best possible accuracy of the all crops baseline.
	
	While PEViD dataset contains relatively simple challenge of low density videos, our 4K dataset of densely populated scenes presents more complicated task. We can see that in the case of the downscale baseline the accuracy is only $33.6 AP_{50}$ while our method achieves $74.3 AP_{50}$, which is very near the performance of the all crops baseline $75.4 AP_{50}$. Our method achieves accuracy as if it was actually inspecting all crops of the original image.
	

	Figure \ref{fig:3_1_num_object_detected} shows the distance between the number of detected objects and the ground truth count. We can see that there is variable density of human figures in each video sequence. We note that in dense scenes we benefit from more detailed setting such as in the case of video “S10” which contains more than 130 human figures. The main factor in different numbers of detected objects due to settings choice are the human figures present in the very distant background of the image. 


\subsection{Performance analysis}

	\begin{figure}[h]
		\centering
		\includegraphics[width=0.5\textwidth]{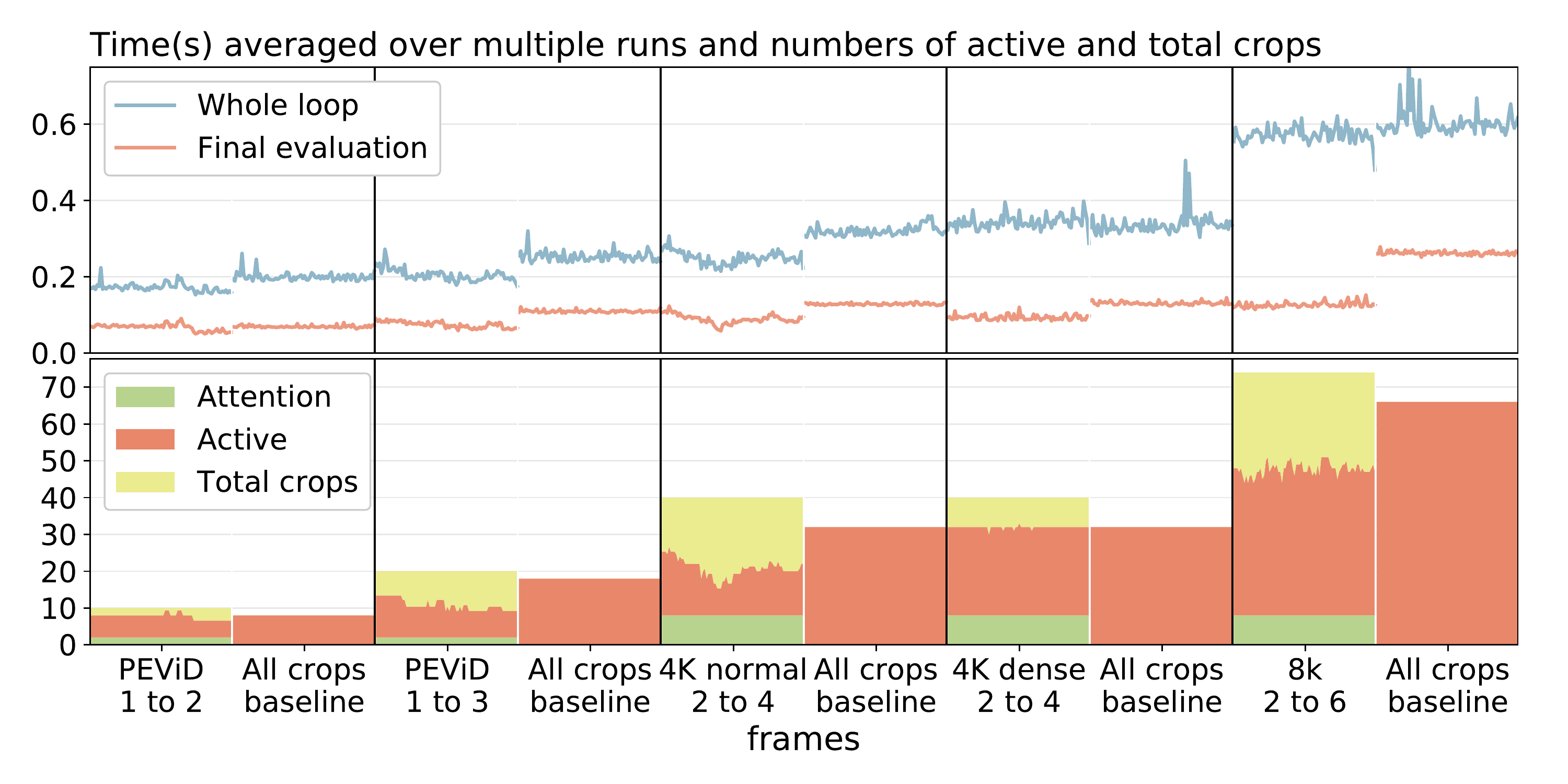}
		\caption{Comparison of the influence of the number of active crops and the setting of resolution of each crop on the speed performance of one frame. Sample of 80 frames is shown for each video sequence.}
		\label{fig:3_2_active_crops_and_speed}
	\end{figure}

	In Figure \ref{fig:3_2_active_crops_and_speed} we see the visualization of amount of active and total possible crops and its influence on processing speed. We also present comparison between our method and the all crops baseline approach.
	
\begin{figure}[h]
		\centering
		\includegraphics[width=0.5\textwidth]{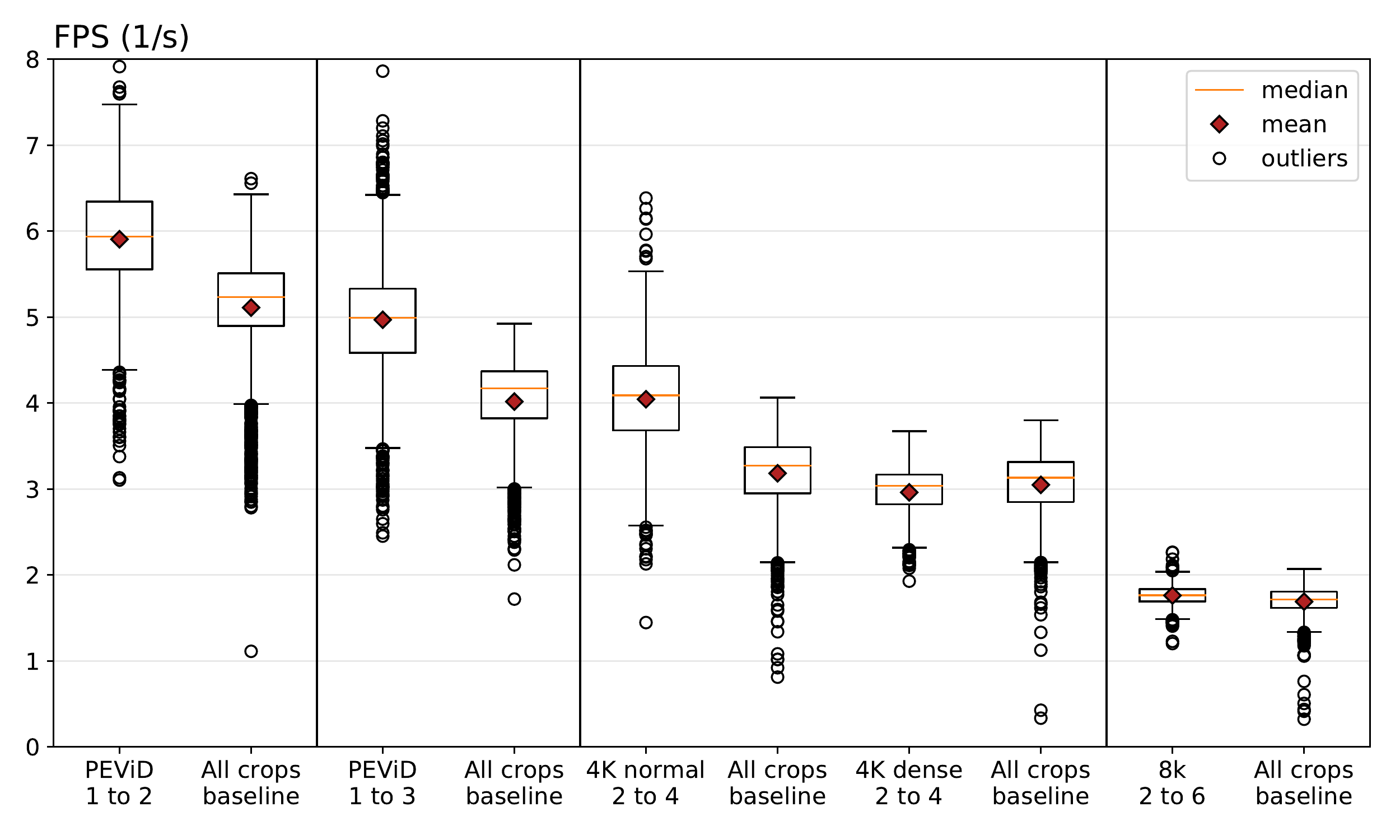}
		\caption{FPS analysis across multiple runs using the best setting.}
		\label{fig:3_2_fps}
	\end{figure}

	In Figure \ref{fig:3_2_fps} we compare the FPS performance of our attention pipeline model with the all crops baseline approach. We note that on an average video from the PEViD dataset our method achieves average performance of 5-6 fps. Scenes which require higher level of detail range between 3-4 fps depending on the specific density. On a bigger and more complex scene of 8K video we achieve 2 fps. Except for the very dense 4K video, our method outperforms the baseline approach.

\begin{figure}[h]
		\centering
		\includegraphics[width=0.4\textwidth]{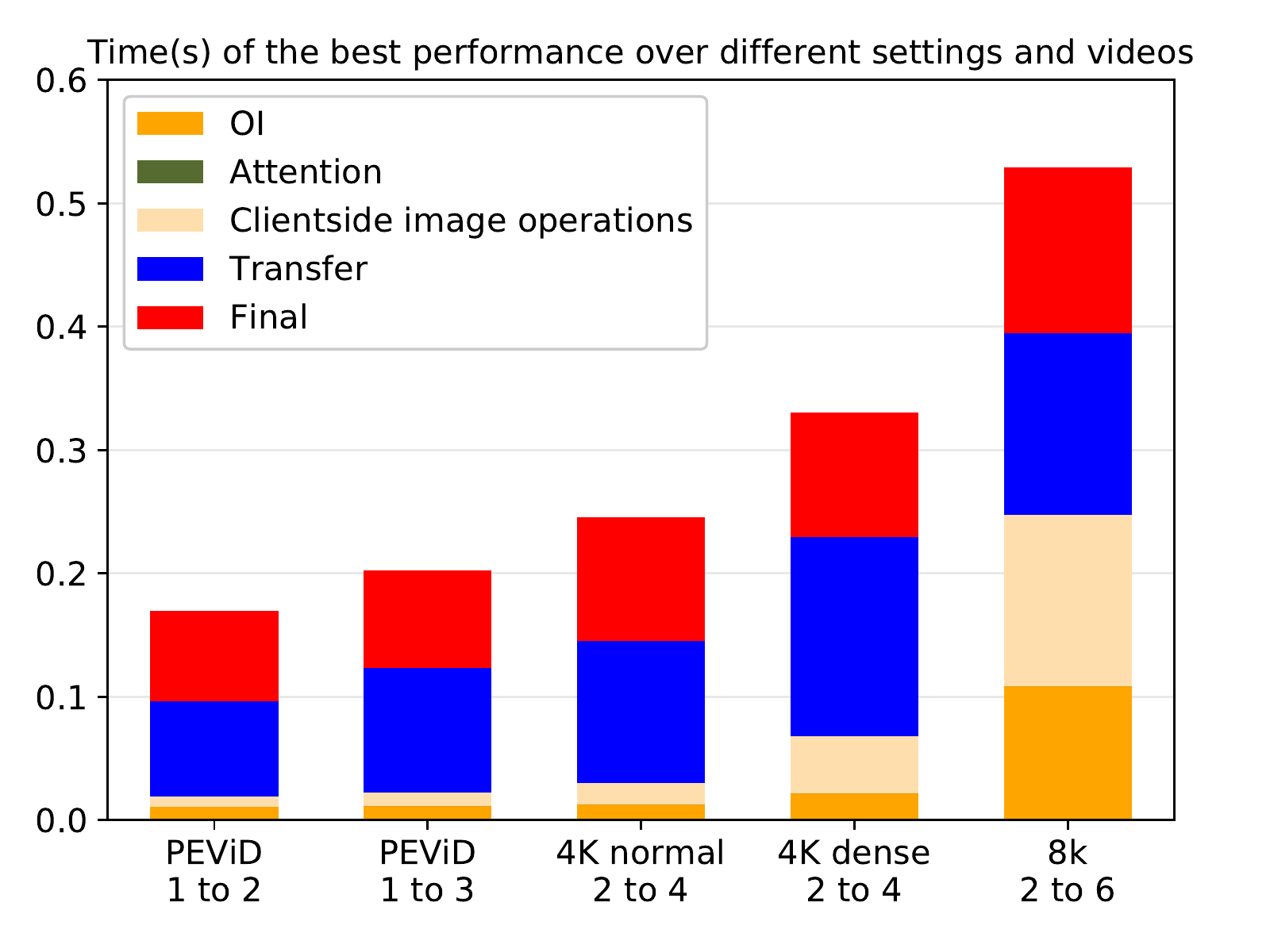}
		\caption{Comparison of average run times on different datasets under their best performance.}
		\label{fig:3_2_best_perf_stacked}
	\end{figure}

	Upon inspecting the detailed decomposition of operations performed in each frame in Figure \ref{fig:3_2_best_perf_stacked}, we can see that the final evaluation in often not the most time consuming step. We need to consider client side operations and the transfer time between one client and many used servers. Note that attention evaluation stays negligible as we are using additional servers for concurrent computation of the next frame. In the case of 8K videos, the I/O time of opening and saving an image becomes a concern as well even as it is performed on another thread. 


	\begin{figure}[h]
		\centering
		\includegraphics[width=0.4\textwidth]{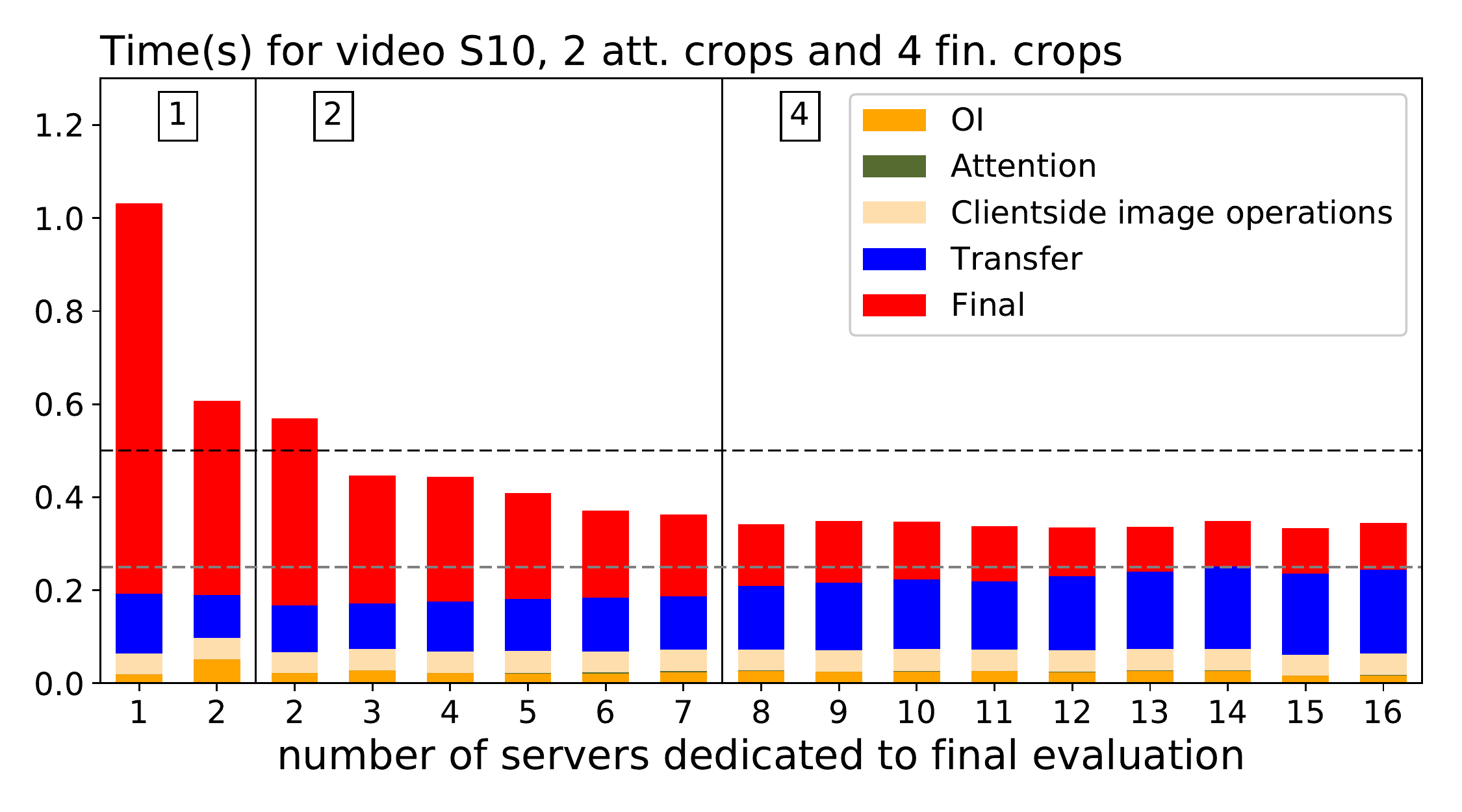}
		\caption{Measured per frame processing speed on our custom 4K video named "S10" with variable amount of servers. The numbers above the chart indicate the amount of servers dedicated for attention precomputing.}
		\label{fig:3_2_S10video}
	\end{figure}

    \if False
	\begin{figure}[h]
		\centering
		\includegraphics[width=0.45\textwidth]{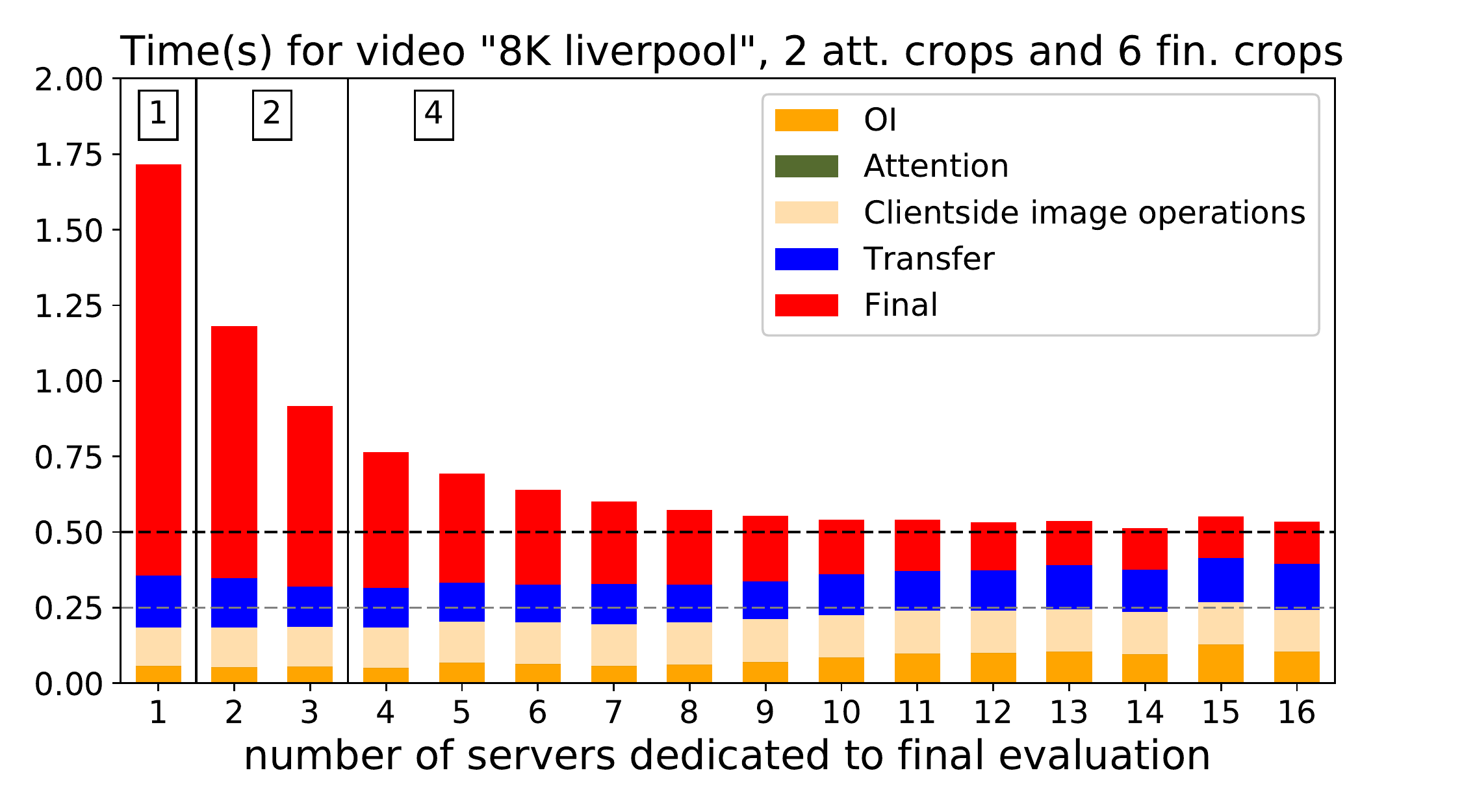}
		\caption{Measured per frame processing speed on downloaded 8K video sample with variable amount of servers. The numbers above the chart indicate the amount of servers dedicated for attention precomputing.}
		\label{fig:3_2_8Kvideo}
	\end{figure}
	\fi

	Finally, we have explored the influence of number of used server for attention precomputation stage and for the final evaluation stage in Figure \ref{fig:3_2_S10video}. We can see, that there is a moment of saturation in scaling the number of workers. This is due to finite amount of crops generated in each frame - after certain number of crops assigned to each server, we don’t see any speedups in further division.



	



\section{Conclusion}

	As a motivation of this paper we have stated two goals in processing high resolution data. First goal consists of the ability to detect even small details included in the 4K or 8K image and not loosing them due to downscaling. Secondly we wanted to achieve fast performance and save on the number of processed crops as compared with the baseline approach of processing every crop in each frame.
	
	Our results show that we outperform the individual baseline approaches, while allowing the user to set the desired trade-off between accuracy and performance.


\ifCLASSOPTIONcompsoc
  \section*{Acknowledgments}
\else
  \section*{Acknowledgment}
\fi


This work used the Extreme Science and Engineering Discovery Environment (XSEDE), which is supported by National Science Foundation grant number 1548562. Specifically, it used the Bridges system, which is supported by NSF award number 1445606, at the Pittsburgh Supercomputing Center (PSC). This work was also sponsored DARPA BRASS program under agreement FA8750-16-2-003 and DARPA PERFECT program under agreement HR0011-13-2-0007.


\bibliographystyle{unsrt}
\bibliography{main}

%
%
%

\end{document}